\pdfoutput=1

\documentclass[11pt]{article}

\usepackage[preprint]{acl}

\usepackage{times}
\usepackage{latexsym}

\usepackage[T1]{fontenc}

\usepackage[utf8]{inputenc}

\usepackage{microtype}

\usepackage{inconsolata}

\usepackage{graphicx}
\usepackage{amssymb}
\usepackage{hyperref}       
\usepackage{url}            
\usepackage{booktabs}       
\usepackage{amsfonts}       
\usepackage{amsmath}  
\usepackage{nicefrac}       
\usepackage{microtype}      
\usepackage{xcolor}         
\usepackage{bm}
\usepackage{graphicx}
\usepackage{subcaption}
\usepackage{cleveref}
\usepackage{listings}

\let\v\bm

\title{Train One Sparse Autoencoder Across Multiple Sparsity Budgets to Preserve Interpretability and Accuracy}

\author{
 \textbf{Nikita Balagansky\thanks{Corresponding author: n.n.balaganskiy@tbank.ru}\textsuperscript{$\clubsuit$,$\heartsuit$ }},
 \textbf{Yaroslav Aksenov\textsuperscript{$\clubsuit$}},
 \textbf{Daniil Laptev\textsuperscript{$\clubsuit$,$\heartsuit$}},
 \textbf{Vadim Kurochkin\textsuperscript{$\clubsuit$,$\heartsuit$}},
\\
 \textbf{Gleb Gerasimov\textsuperscript{$\clubsuit$,$\heartsuit$,$\spadesuit$}},
 \textbf{Nikita Koriagin\textsuperscript{$\clubsuit$}},
 \textbf{Daniil Gavrilov\textsuperscript{$\clubsuit$}}\\
 \textsuperscript{$\clubsuit$}T-Tech,
 \textsuperscript{$\heartsuit$}Moscow Institute of Physics and Technology
 \textsuperscript{$\spadesuit$}HSE University
}

\begin{document}
\maketitle
\begin{abstract}
Sparse Autoencoders (SAEs) have proven to be powerful tools for interpreting neural networks by decomposing hidden representations into disentangled, interpretable features via sparsity constraints. However, conventional SAEs are constrained by the fixed sparsity level chosen during training; meeting different sparsity requirements therefore demands separate models and increases the computational footprint during both training and evaluation. We introduce a novel training objective, \emph{HierarchicalTopK}, which trains a single SAE to optimise reconstructions across multiple sparsity levels simultaneously. Experiments with Gemma-2 2B demonstrate that our approach achieves Pareto-optimal trade-offs between sparsity and explained variance, outperforming traditional SAEs trained at individual sparsity levels. Further analysis shows that HierarchicalTopK preserves high interpretability scores even at higher sparsity. The proposed objective thus closes an important gap between flexibility and interpretability in SAE design.

\end{abstract}

\section{Introduction}

Transformers have revolutionised natural language processing (NLP) by achieving state-of-the-art performance across diverse tasks. Yet their internal representations remain notoriously difficult to interpret, often exhibiting \emph{polysemanticity}, in which individual neurons activate for semantically unrelated features. To address this challenge, recent work has focused on Sparse Autoencoders (SAEs), which learn disentangled, human-interpretable directions in Transformer residual streams by enforcing sparsity constraints on the latent representations.

SAEs decompose hidden states into latent embeddings that are theoretically grounded in the independent additivity principle \citep{ayonrinde2024compression}. This principle posits that individual features contribute to model behaviour independently, enabling isolated analysis of the latents. In practice, relaxing sparsity constraints (e.g.\ increasing the number of active latents) often introduces entanglement: latents begin to co-activate for unrelated features, undermining interpretability. Consequently, the effectiveness of existing SAEs is tightly coupled to a single sparsity level fixed during training.

We propose \emph{HierarchicalTopK}, a novel activation mechanism and training objective that enables a single SAE to maintain interpretable features across a range of sparsity levels. Unlike conventional SAEs, which must be retrained to accommodate different sparsity requirements, our method ensures that any subset of latents with $k\le K$ remains disentangled and faithful to the independent additivity principle. Empirically, HierarchicalTopK SAEs achieve Pareto-optimal trade-offs between sparsity and explained variance, outperforming traditional SAEs trained independently at varying sparsity levels. This work bridges the gap between flexibility and interpretability in SAE design, enabling dynamic adaptation to downstream tasks with varying computational or fidelity requirements.


\section{Method}
\label{sec:method}

\begin{figure}[hbtp]
    \centering
    \includegraphics[width=0.99\linewidth]{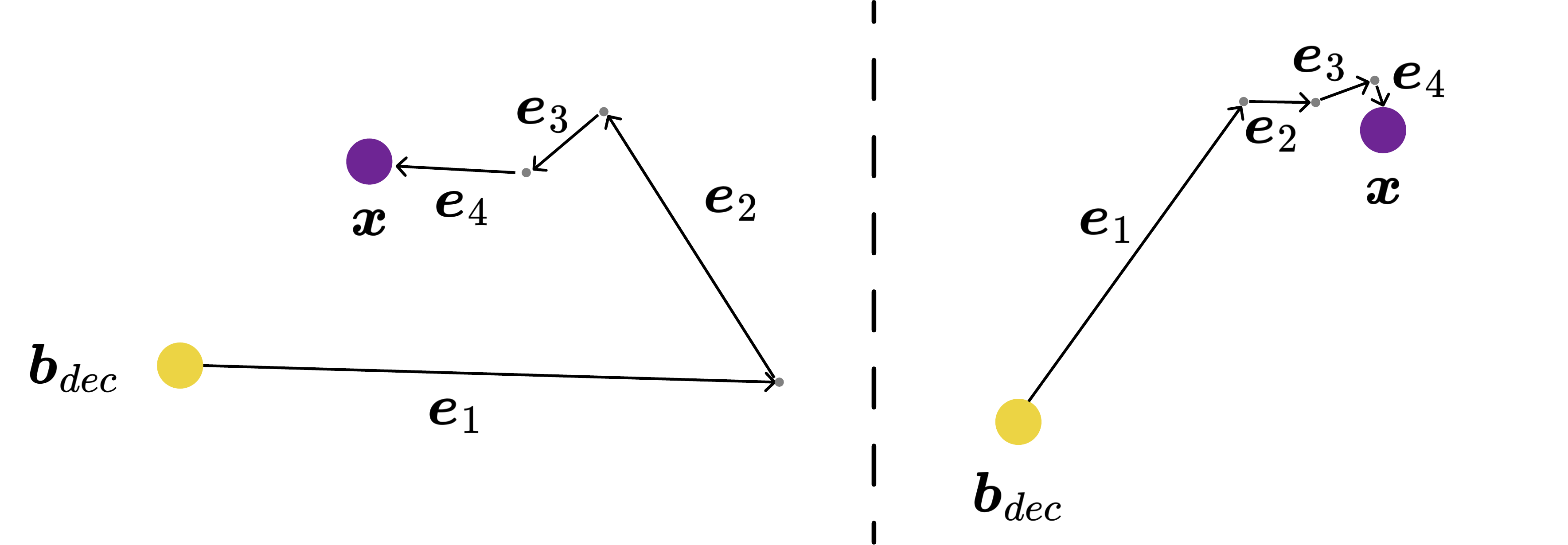}
    \caption{\textbf{Left:} SAE trained on a single $k$. \textbf{Right:} SAE trained on all $k \le K$.}
    \label{fig:scheme}
\end{figure}

A sparse autoencoder (SAE) is defined as
\begin{align*}
        \v{l} &= \sigma\!\bigl(W_{\text{enc}}\v{x} + \v{b}_{\text{enc}}\bigr),\\
        \hat{\v{x}} &= W_{\text{dec}}\v{l} + \v{b}_{\text{dec}},
\end{align*}
where $W_{\text{enc}}\!\in\!\mathbb{R}^{D\times h}$, $W_{\text{dec}}\!\in\!\mathbb{R}^{h\times D}$, $\v{b}_{\text{enc}}\!\in\!\mathbb{R}^{D}$, and $\v{b}_{\text{dec}}\!\in\!\mathbb{R}^{h}$. Here, $D$ is the dictionary size and $h$ the hidden dimension.  
The non-linearity $\sigma(\cdot)$ is central.  
Vanilla SAEs use $\operatorname{ReLU}$ \citep{bricken2023monosemanticity}, requiring an additional sparsity penalty on the latents.  
Sparsity can instead be induced directly with activations such as \textsc{TopK} \citep{makhzani2013ksparse} or \textsc{BatchTopK} \citep{bussmann2024batchtopk}. Our analysis focuses on these activation variants.

The decoder can be viewed as a set of embeddings $W_{\text{dec}}=[\v{e}_1,\ldots,\v{e}_D]$, yielding
\begin{equation*}
    \hat{\v{x}} = \sum_{i\in\operatorname{top}_k}\! \v{l}_i(\v{x})\,\v{e}_i + \v{b}_{\text{dec}},
\end{equation*}
where $\v{l}_i$ is the $i$-th component of $\v{l}$.  
Embeddings are thus scaled by $\v{l}_i(\v{x})$ to reconstruct $\v{x}$.  
The reconstruction error is $\mathcal{L}_{\text{rec}} = \|\v{x} - \hat{\v{x}}\|^2$.  
Optimising $\mathcal{L}_{\text{rec}}$ for a fixed $\operatorname{top}_k$ can be suboptimal when one wishes to interpret individual directions (small $k$).

\paragraph{Hierarchical loss.}
We therefore introduce a \emph{hierarchical} loss. Define
\begin{align*}
    \hat{\v{x}}_j &= \sum_{i\in\operatorname{top}_j}\! \v{l}_i(\v{x})\,\v{e}_i + \v{b}_{\text{dec}},\\
    \mathcal{L}^{j}_{\text{rec}} &= \|\v{x} - \hat{\v{x}}_j\|^2,
\end{align*}
for $j \in \mathcal{J} \subset \mathbb{N}$ (e.g.\ $\mathcal{J}=\{1,\dots,k\}$).  
The overall objective is
\begin{equation}
\label{eq:loss}
    \mathcal{L}_{\text{hierarchical}} = \frac{1}{|\mathcal{J}|}\sum_{j\in\mathcal{J}}\mathcal{L}^{j}_{\text{rec}}.
\end{equation}

Whereas the standard SAE guarantees reconstruction only at $k$ active embeddings, our formulation encourages good reconstructions for every $j\le k$.  
The optimal model under $\mathcal{L}_{\text{hierarchical}}$ therefore improves $\hat{\v{x}}_j$ monotonically with increasing $j$, a property absent in the vanilla SAE.

The hierarchical loss is inexpensive: it can be computed in a single forward pass via a cumulative-sum operation and implemented with kernels that avoid materialising intermediate tensors. In our implementation it runs faster than the original TopK loss; see Appendix~\ref{ap:implementation} for details.

\section{Experiments}

\subsection{Setup}

For our experiments, we chose the Gemma-2 2B model \citep{team2024gemma}. We trained SAEs on a 1 B-token subsample of the FineWeb dataset \citep{penedo2024fineweb}. Unless stated otherwise, we use the output of the 12th Transformer layer and set the SAE dictionary size to $D = 65\,536$. Training details are provided in Appendix~\ref{sec:appendix_training_details}.

We report the fraction of unexplained variance (FVU) as the main metric:
\begin{equation*}
    \operatorname{FVU}(\v{x}, \v{\hat{x}}) = \frac{\operatorname{Var}(\v{x} - \v{\hat{x}})}{\operatorname{Var}(\v{x})}.
\end{equation*}

Sparsity is measured by the $\ell_0$ norm
\begin{equation*}
    \ell_0 = \sum_i \mathbf{I}[\v{l}_i >0].
\end{equation*}
Because we use TopK-based activations, $\ell_0 = k$.


         

\subsection{Hierarchical SAE Pareto Frontier}

\begin{figure}
    \centering
    \includegraphics[width=0.7\linewidth]{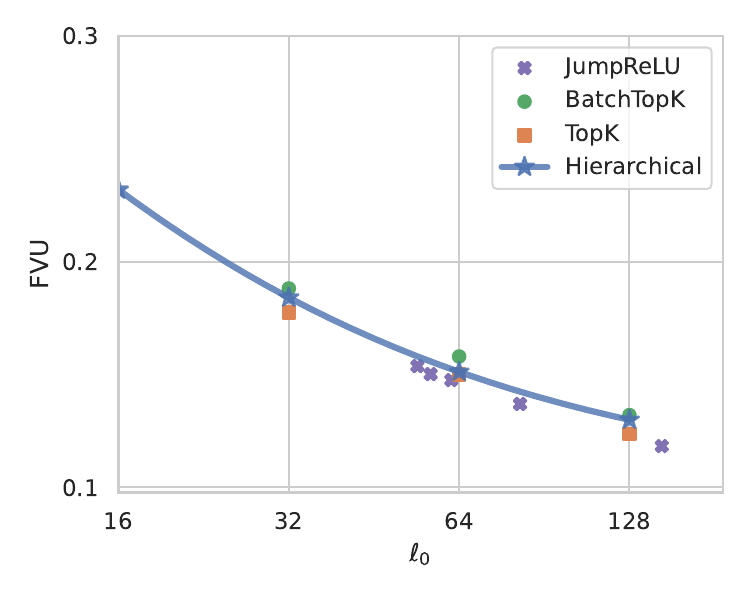}
    \caption{Comparison of an SAE with Hierarchical activation against other activation variants. The proposed method lies on the Pareto-optimal frontier across all sparsity levels, even though it is a single model.}
    \label{fig:scatter}
\end{figure}

To evaluate the proposed training technique, we trained baseline SAEs at different sparsity levels. Specifically, we trained JumpReLU~\citep{rajamanoharan2024jumpingahead} with various sparsity-regularisation coefficients and TopK and BatchTopK SAEs with $k \in \{32, 64, 128\}$. We also trained a single HierarchicalTopK SAE with $K = 128$. Figure~\ref{fig:scatter} shows that our model matches or surpasses the performance of the individually trained baselines across all sparsity levels while requiring only one set of parameters.

\subsection{Changing $\ell_0$ at the Inference}
\label{sec:l0_change}

\begin{figure*}[htbp]
    \centering
    \begin{subfigure}{0.35\textwidth}
        \centering
        \includegraphics[width=\linewidth]{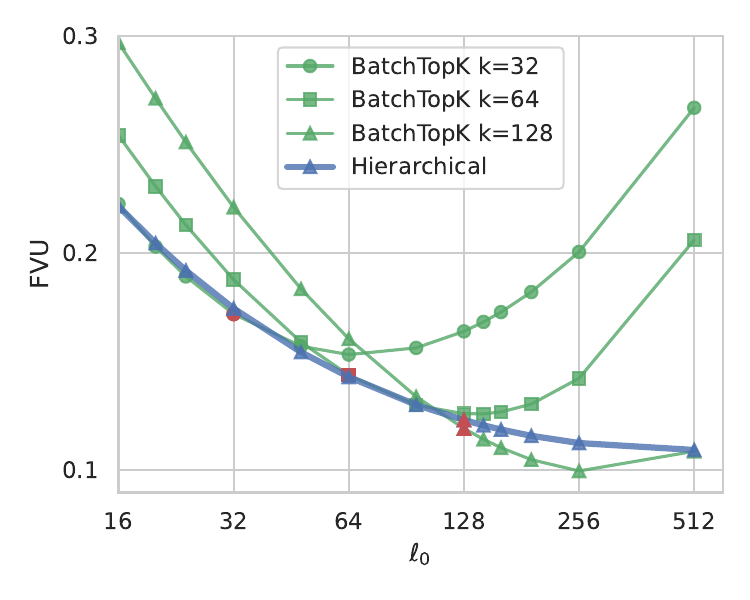}
        \caption{BatchTopK}
        \label{subfig:dict_size_16384}
    \end{subfigure}
    \begin{subfigure}{0.35\textwidth}
        \centering
        \includegraphics[width=\linewidth]{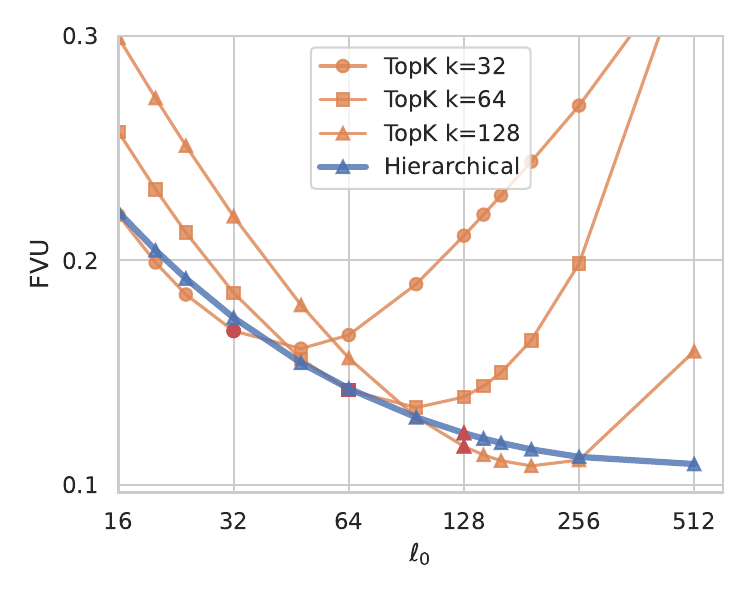}
        \caption{TopK}
        \label{subfig:dict_size_32768}
    \end{subfigure}
    \caption{Pareto frontier for SAEs with BatchTopK, TopK, and Hierarchical activation functions. Red dots denote the $\ell_0$ values on which the BatchTopK and TopK SAEs were trained. HierarchicalTopK matches or surpasses separately trained BatchTopK and TopK SAEs when interpolating ($\ell_0 \le 128$), allowing a single SAE to select $\ell_0$ post-training. See Section~\ref{sec:l0_change} for details.}

    \label{fig:main}
\end{figure*}

To assess generalisation across sparsity levels, we trained a single HierarchicalTopK SAE with $K = 128$ and baseline TopK and BatchTopK SAEs with fixed $k \in \{32, 64, 128\}$. At inference we varied $\ell_0$ over a dense grid, including both interpolation points within the training range and extrapolation points outside it. As shown in Figure~\ref{fig:main}, the Hierarchical model performs as well as—or better than—the baselines for $\ell_0 \le 128$, demonstrating that training across multiple $k$ values is crucial for robust performance.

BatchTopK mixes different $k$ values between samples during training, resulting in a primitive form of extrapolation. Consequently, it continues to improve reconstructions for $\ell_0 \in [128, 512]$, a sparsity range rarely used in practice.

\subsection{Pointwise Loss}
\label{sec:pointwise_loss}

\begin{figure}[hbtp]
    \centering
    \includegraphics[width=0.7\linewidth]{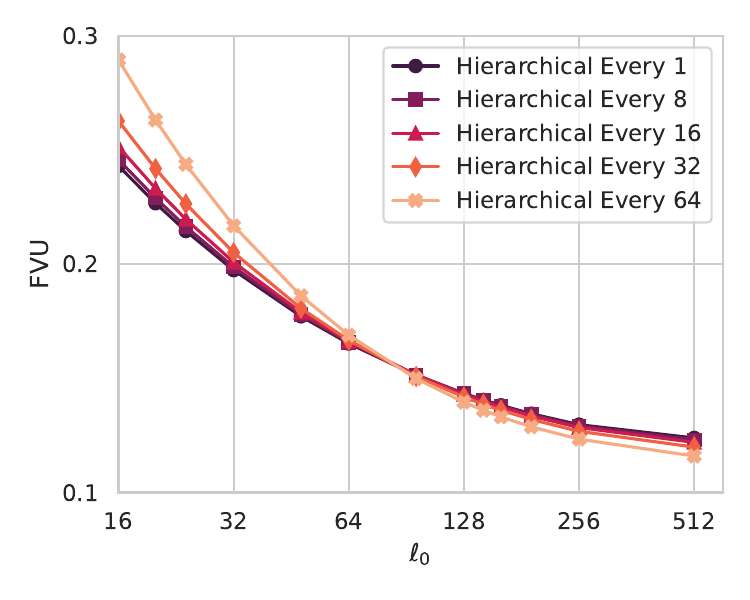}
    \caption{We test simple heuristics to reduce the computation required to train HierarchicalTopK. Computing the loss on every 8th term does not affect performance; see Section~\ref{sec:pointwise_loss} for details.}
    \label{fig:pointwise_loss}
\end{figure}

To reduce computational overhead we evaluated computing the hierarchical loss on a subsampled index set (Equation~\ref{eq:loss}):
\begin{equation*}
J_x = \{1\} \cup \{i \in \mathbb{N} : i \bmod x = 0 \land 1 < i \le k \},    
\end{equation*}
with $x \in \{1, 8, 16, 32, 64\}$. As Figure~\ref{fig:pointwise_loss} shows, computing the loss on every 8th term ($x = 8$) yields performance indistinguishable from the full loss, providing an eight-fold theoretical reduction in FLOPs.

\begin{figure}[hbtp]
    \centering
    \includegraphics[width=0.7\linewidth]{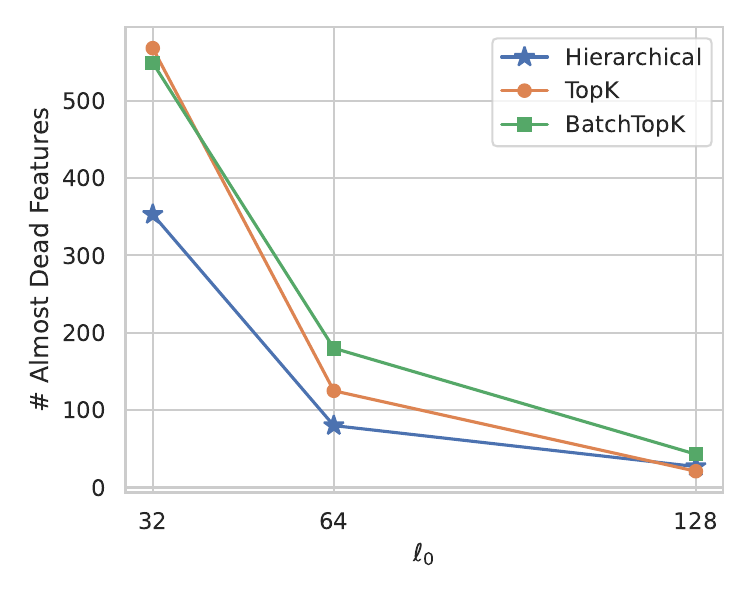}
    \caption{Number of features with activation frequency below $10^{-5}$ (``almost dead'') for SAEs trained with $k=128$. ``Optimal scaling'' denotes the number of almost-dead features in a BatchTopK SAE trained with $k = \ell_0$. The BatchTopK model accumulates almost-dead features more rapidly than the Hierarchical model when $k$ is reduced at inference time; see Section~\ref{sec:justification} for details.}
    \label{fig:dead_features}
\end{figure}

An SAE whose loss is calculated on every 64th term suffers a significant performance decrease for $\ell_0 < 128$, but extrapolates better for $\ell_0 > 128$. Remarkably, using the hierarchical loss on every 8th term ($J_8 = \{1, 8, 16, 24, 32, \dots, 128\}$) reduces theoretical overhead by a factor of eight without sacrificing reconstruction quality. In practice, however, there is almost no difference in per-step training time between vanilla TopK and HierarchicalTopK; see Appendix~\ref{ap:implementation} for details.

\begin{figure*}[htbp]
    \centering
    \begin{subfigure}{0.35\textwidth}
        \centering
        \includegraphics[width=\linewidth]{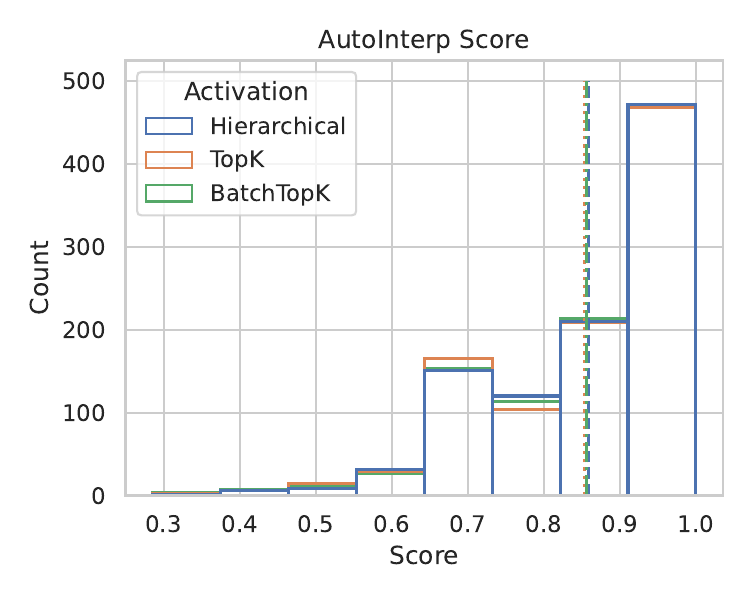}
        \caption{$\ell_0=32$}
        \label{subfig:interp_32}
    \end{subfigure}
    \begin{subfigure}{0.35\textwidth}
        \centering
        \includegraphics[width=\linewidth]{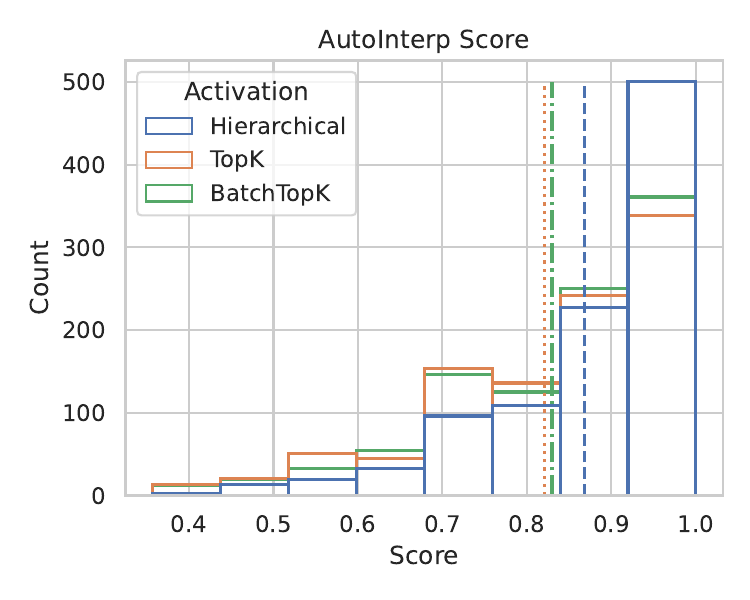}
        \caption{$\ell_0 = 128$}
        \label{subfig:interp_128}
    \end{subfigure}
    \caption{AutoInterp Score \citep{paulo2024automatically}. TopK and BatchTopK scores are obtained from two separate SAEs trained with $k = 32$ and $k = 128$; the Hierarchical model uses a single SAE trained on all $k \le 128$. Hierarchical activation preserves the interpretability level of SAEs trained with smaller $\ell_0$.}
    \label{fig:interp}
\end{figure*}

\subsection{Why SAE Struggle to Reduce $\ell_0$?}
\label{sec:justification}

To investigate why simple SAE variants struggle to interpolate to lower $\ell_0$ values than those used during training, we measured the number of features whose activation frequency falls below $10^{-5}$ (i.e.\ they activate once in $10^{5}$ tokens). We call these features \emph{almost dead}.

We trained TopK and BatchTopK models with $k = 128$ and then, following Section~\ref{sec:l0_change}, evaluated them with $k \in \{32, 64, 128\}$. The Hierarchical variant was trained once with $k = 128$. Results are shown in Figure~\ref{fig:dead_features}. Although all SAEs exhibit similar numbers of dead features at $\ell_0 = 128$, the Hierarchical model keeps significantly more features alive than the TopK and BatchTopK variants as $k$ decreases.


\subsection{Interpretability}

To validate interpretability we use the detection score of \citet{paulo2024automatically}, implemented in SAE Bench \citep{karvonen2025saebench}. For TopK and BatchTopK we evaluate two SAEs trained with $k=32$ and $k=128$; for Hierarchical we evaluate a single SAE trained on all $k\le 128$ (see Figure~\ref{fig:interp}).

For the Hierarchical SAE the interpretability score at $\ell_0 = 128$ is almost identical to that at $\ell_0 = 32$, while its explained variance remains on the Pareto frontier (Figure~\ref{fig:scatter}). By contrast, in both TopK and BatchTopK variants, less-sparse models tend to be less interpretable. This observation underscores the superiority of the hierarchical loss.

\section{Related Work}

Research on sparse autoencoders increasingly focuses on feature interpretability in Transformer representations. The seminal work of \citet{gao2025scaling} introduced TopK-sparse autoencoders; \citet{bussmann2024batchtopk} extended this idea with batch-level sparsity control. However, these approaches lack a mechanism for establishing feature importance or relationships.

Structural constraints have also been explored. \citet{bussmann2025matryoshka} and \citet{ayonrinde2024compression} investigate hierarchical dictionaries, demonstrating the benefits of progressive refinement. Building on these insights, our training method naturally encodes feature importance through progressive reconstruction, mirroring gradient-descent dynamics and feature hierarchies while maintaining interpretability and improving generalisation across sparsity levels.

\section{Conclusion}
We introduced \emph{HierarchicalTopK}, a single sparse-autoencoder objective that enforces high-quality reconstructions at every sparsity level up to a chosen budget $K$. Experiments on Gemma-2 2B representations show that our approach:
\begin{itemize}
  \item Achieves Pareto-optimal trade-offs between explained variance and $\ell_0$ compared with independently trained TopK and BatchTopK baselines, despite using a single model.
  \item Maintains high interpretability across sparsity levels and prevents the proliferation of ``dead’’ features when $\ell_0$ is varied at inference time.
\end{itemize}
These contributions provide a flexible, efficient, and interpretable framework for analysing Transformer latent spaces under varying computational constraints.

\section{Limitations}

Our work has two principal limitations. (\textit{i}) \textbf{Evaluation scope}: experiments are limited to the Gemma-2 2B model and a FineWeb subset; transfer to other architectures and datasets remains to be tested. (\textit{ii}) \textbf{Interpretability measures}: we rely on automated metrics as proxies for human judgement; user studies are needed to validate semantic alignment.

\bibliography{custom}

\appendix
\onecolumn

\section{SAE Training Details}
\label{sec:appendix_training_details}

All SAEs were trained with a modified version of the code from \citet{bussmann2024batchtopk}. Hyperparameters are listed in Table~\ref{tab:hyperparameters}. We use NVIDIA H100 80GB GPU and spent about 20 GPU-days of compute, including preliminary experiments.

\begin{table}[hbtp]
    \centering
    \begin{tabular}{r|c}
    \toprule
        Parameter & Value \\ \midrule
        Optimizer & Adam \\
        Learning Rate & $0.0008$ \\
        \# tokens & $10^9$ \\
        Dataset & FineWeb \\
        Batch Size & $8096$ \\ 
        Decoder Normalization & True \\\bottomrule
    \end{tabular}
    \caption{Hyperparameters used to train the SAEs. See Section \ref{sec:appendix_training_details} for more details.}
    \label{tab:hyperparameters}
\end{table}

We also used modified kernels from \citet{gao2025scaling}; see Appendix~\ref{ap:implementation} for details.

\section{Additional Results}

\subsection{Latent Structure}

To support Figure~\ref{fig:scheme} we measured the cosine similarity between feature embeddings in the reconstruction sum

\begin{equation*}
    \hat{\v{x}} = \sum_{i\in\operatorname{top}_k}\! \v{l}_i(\v{x})\,\v{e}_i + \v{b}_{\text{dec}},
\end{equation*}

\begin{figure}[htbp]
    \centering
    \includegraphics[width=0.4\linewidth]{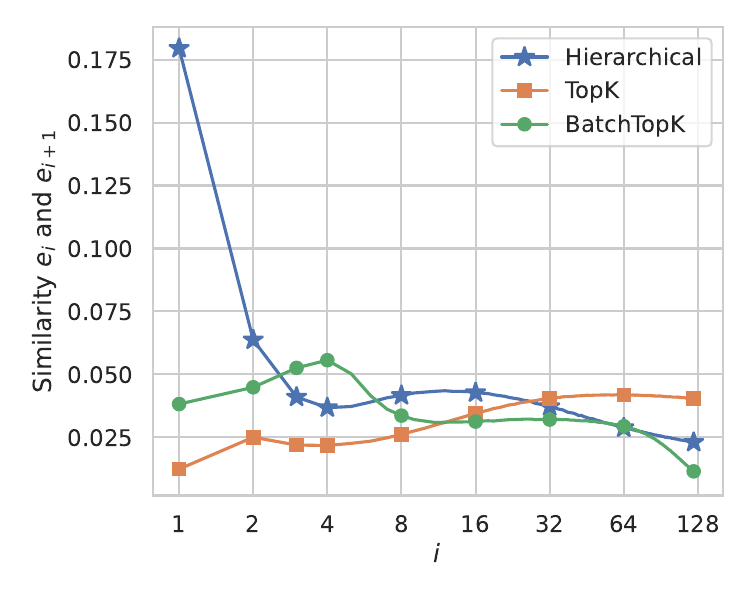}
    \caption{Cosine similarity of feature embeddings in the reconstruction sum.}
    \label{fig:similarity}
\end{figure}

In Figure~\ref{fig:similarity}, $e_1$ denotes the top-1 activation, and so on. Ideally, similarity should decrease monotonically as activation values diminish.

Vanilla TopK SAEs show the undesired trend that similarity increases with the index $i$, whereas the Hierarchical model preserves the expected monotonic decrease.

\subsection{Distribution of the Latents Activations}

\begin{figure}[htbp]
    \centering
    \begin{subfigure}{0.4\textwidth}
        \includegraphics[width=\linewidth]{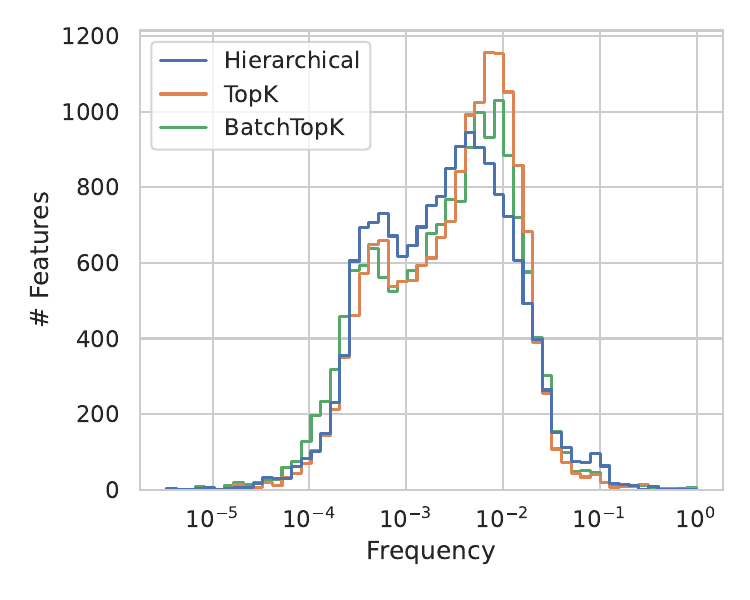}
        \caption{Feature frequency}
    \end{subfigure}
    \begin{subfigure}{0.4\textwidth}
        \includegraphics[width=\linewidth]{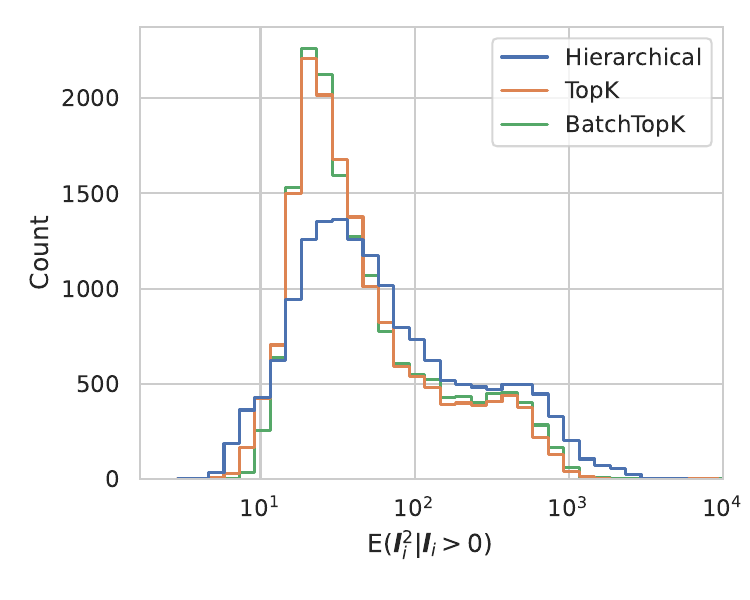}
        \caption{Mean squared activation}
        \label{subfig:squared_acts}
    \end{subfigure}
    \caption{Latent‐feature distributions for SAEs trained with $k = 128$ ($J=\{1,\dots,k\}$ in Hierarchical training).}
    \label{fig:distributions}
\end{figure}

To compare the distributions learned by standard SAEs and the Hierarchical variant we analyse both feature frequency and mean-squared activation (Figure~\ref{fig:distributions}).  
Hierarchical training yields more latents with higher activation values (panel~\ref{subfig:squared_acts}), which may explain its superior interpretability.  
Its frequency distribution is skewed towards lower values, indicating that the hierarchical loss encourages activations to appear as the top-1 feature, enabling accurate reconstruction even at $k = 1$.

\subsection{JumpReLU and TopK evaluations}

\begin{figure}[hbtp]
    \centering
    \begin{subfigure}{0.49\linewidth}
        \includegraphics[width=\linewidth]{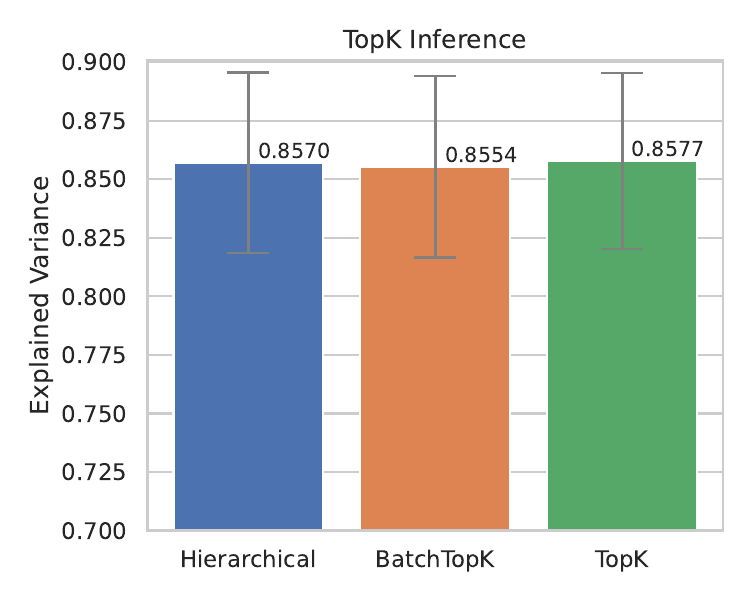}
    \end{subfigure}
    \begin{subfigure}{0.49\linewidth}
        \includegraphics[width=\linewidth]{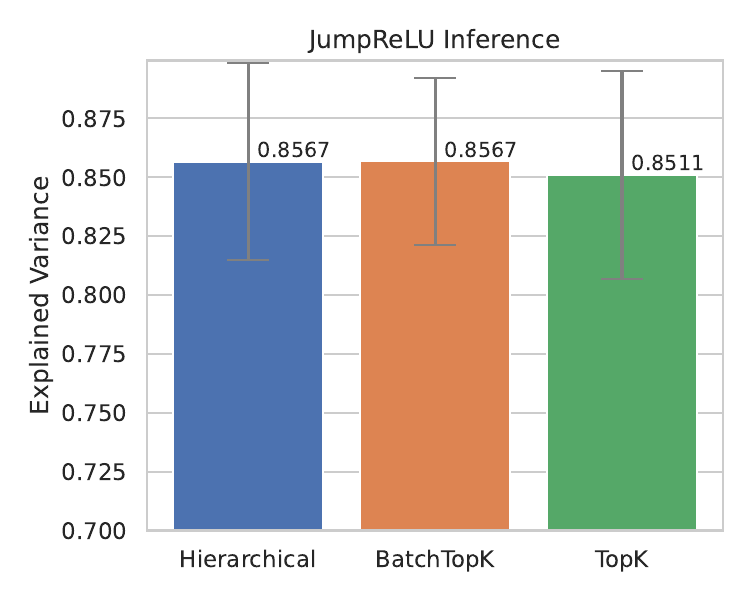}
    \end{subfigure}
    \caption{TopK versus JumpReLU inference. We did not find a significant difference between JumpReLU and fixed TopK evaluation.}
    \label{fig:jumprelu}
\end{figure}

A BatchTopK SAE is trained with batch-wise sparsity but, if evaluated directly with per-token TopK, the training and inference settings mismatch.  
We therefore apply a constant-threshold JumpReLU at inference time, choosing the threshold so that the expected number of active features equals $k$.  
To study the effect of switching activations we trained SAEs with $\ell_0 = 64$ (the Hierarchical model was trained on $k \le 128$) and evaluated every model at $\ell_0 = 64$.  
Results are shown in Figure~\ref{fig:jumprelu}.

The largest change in explained variance occurs for the TopK SAE, which drops from $0.8577$ to $0.8511$ under JumpReLU.  
BatchTopK improves marginally ($+0.0013$), and the Hierarchical variant is virtually unchanged for either activation.

\section{Implementation}
\label{ap:implementation}

We extend the Triton implementation \citep{triton}\footnote{\url{https://github.com/openai/sparse_autoencoder}} of TopK-sparse autoencoders by fusing mean-squared-error computation directly into the sparse-decoder kernel.  
This produces both our optimised \textit{TopKSAE} (with a fused loss) and \textit{FlexSAE}.  
As shown in Table~\ref{tab:kernels}, both kernels surpass the baseline Triton implementation in speed while using the same peak memory.

In addition, we provide a minimal PyTorch implementation of the FlexSAE loss for clarity.  This naïve version illustrates the core idea: we gather the decoder embeddings at the active indices, scale them by the sparse activations, compute a cumulative reconstruction, and measure the mean-squared error across all sparsity levels.

\begin{lstlisting}[language=Python, caption={Naïve PyTorch implementation of the Hierarchical loss.}]
def hierarchical_loss(sparse_idx, sparse_val, decoder, b_dec, target):
    """
    sparse_idx: LongTensor of shape (B, K) with indices of active embeddings
    sparse_val: FloatTensor of shape (B, K) with corresponding activation values
    decoder:     FloatTensor of shape (D, h) containing the dictionary embeddings
    b_dec:       FloatTensor of shape (h) containing decoder bias
    target:      FloatTensor of shape (B, h) with the original inputs
    """
    B, K = sparse_idx.shape
    flatten_idx = sparse_idx.view(-1)
    emb = decoder[flatten_idx].view(B, K, -1)               
    emb = emb * sparse_val.unsqueeze(-1)  

    recon_cum = emb.cumsum(dim=1) + b_dec.unsqueeze(1)           

    diff = recon_cum - target.unsqueeze(1)
    total_err = diff.pow(2).mean()
    return total_err
\end{lstlisting}

\begin{table}[t]
  \centering
  \begin{tabular}{lcc}
    \toprule
    SAE Method               & Time per step (ms) & Peak Memory (MB) \\
    \midrule
    Topk \citep{gao2025scaling}            & $10.4800 \pm 0.0836$ & 6369.86 \\ \midrule
    Fused Topk (Ours)       & $\mathbf{9.9235 \pm 0.0368}$ (–5.31\%) & 6371.06 \\
    Fused Hierarchical (Ours)           & $\mathbf{10.0531 \pm 0.0420}$ (–4.07\%) & 6371.06 \\
    \bottomrule
  \end{tabular}
    \caption{Training speed and memory usage for sparse-autoencoder kernels with batch size $B = 64$, model dimension $h = 2304$, dictionary size $D = 2^{16}$, and sparsity $\ell_0 = 128$.}
  \label{tab:kernels}
\end{table}
\twocolumn


\end{document}